\documentclass{mva_style}
\usepackage{graphicx}

\finalcopy 

\ifx\argmin\undefined\newcommand{\argmin}{\mathop{\rm argmin}\limits}\fi
\ifx\argmax\undefined\newcommand{\argmax}{\mathop{\rm argmax}\limits}\fi
\ifx\bm\undefined\newcommand{\bm}[1]{\mbox{\boldmath{$#1$}}}\fi
\ifx\um\undefined\newcommand{\um}[1]{{\SI{#1}{\micro \metre}}}\fi

\ifx\etal\undefined\newcommand{\etal}{{\it et al. }}\fi
\ifx\ie\undefined\newcommand{\ie}{{\it i.e.}}\fi
\ifx\eg\undefined\newcommand{\eg}{{\it e.g.}}\fi
\ifx\aka\undefined\newcommand{\aka}{{\it a.k.a.}}\fi

\ifx\figref\undefined\newcommand{\figref}[1]{{Fig.\ref{#1}}}\fi
\ifx\tabref\undefined\newcommand{\tabref}[1]{{Table \ref{#1}}}\fi
\ifx\equref\undefined\newcommand{\equref}[1]{Eq.(\ref{#1})}\fi
\ifx\secref\undefined\newcommand{\secref}[1]{Sec.\ref{#1}}\fi
\ifx\subsecref\undefined\newcommand{\subsecref}[1]{Sec.\ref{#1}}\fi

\newcommand{\bnote}[1]{{\color{red} #1 \color{black}}}

\newcommand{\kcut}[1]{}
\newcommand{\jptext}[1]{{\color{blue} \bf #1 \color{black}}}
\newcommand{\snote}[1]{}
\newcommand{\scut}[1]{}
\ifx\pdfoutput\undefined
\else
 \usepackage{CJKutf8}
 \renewcommand{\bnote}[1]{}
 \renewcommand{\jptext}[1]{}
\fi

\usepackage{color}

\graphicspath{{../}{./}{./figure-ba/}}

\begin{document}
\title{Generalization of pixel-wise phase estimation by CNN and\\
improvement of phase-unwrapping by MRF optimization\\
for one-shot 3D scan}

\author{\footnotesize 
  Hiroto Harada\\
\footnotesize   Kyushu University\\
\footnotesize   Fukuoka, Japan\\
   \and
\footnotesize  Michihiro Mikamo\\
\footnotesize   Hiroshima City University\\
\footnotesize   Hiroshima, Japan\\
   \and
\footnotesize   Ryo Furukawa\\
\footnotesize   Kinki University\\
\footnotesize   Osaka, Japan\\
   \and \vspace{-1mm}
\footnotesize  Ryusuke Sagawa\\ \vspace{-2mm}
  	\tiny Advanced Industrial \\ 
   \tiny Science and Technology\\
\footnotesize  Ibaraki, Japan\\
  \and
\footnotesize   Hiroshi Kawasaki\\
\footnotesize   Kyushu University\\
\footnotesize   Fukuoka, Japan\\
}

\maketitle

\section*{\centering Abstract}
\textit{
Active stereo technique using single pattern projection, \aka~one-shot 3D scan, 
have drawn a wide attention from industry, medical purposes, etc.
One severe drawback of one-shot 3D scan is sparse reconstruction.
In addition, since spatial pattern becomes complicated for the 
    purpose of efficient
    embedding, it is easily affected by noise, which results in unstable decoding.
To solve the problems, we propose a pixel-wise interpolation technique for one-shot scan, 
    which is applicable to any types of static pattern if the pattern is regular 
    and periodic. This is achieved by U-net which is pre-trained by CG with efficient data 
    augmentation algorithm.
In the paper, to further overcome the decoding 
    instability, we 
    propose a robust correspondence finding algorithm based on Markov random 
    field (MRF) optimization.
We also propose a shape refinement algorithm based on b-spline and
    Gaussian kernel interpolation using explicitly detected 
    laser curves.
Experiments are conducted to show the effectiveness of the proposed method using 
    real data with strong noises and textures.
}

\section{Introduction}
\label{sec:intro}

One-shot 3D scanning techniques become significantly important, since it can capture moving 
objects like human face or can be equipped with autonomous car, drones, medical 
robots, etc.
One severe drawback of one-shot 3D scan is sparse reconstruction, because 
    positional information of projector coordinate are encoded into spatial 
    pattern. In addition, since spatial pattern becomes complicated for the 
    purpose of efficient
    embedding, it is easily affected by noise, which results in unstable 
    decoding and low accuracy on 3D shape.

To solve the problems, we have proposed a pixel-wise interpolation technique for one-shot scan, 
    which is applicable to any types of static pattern if the pattern is regular 
    and periodic~\cite{Furukawa_2022_WACV}. This is achieved by U-net which is pre-trained by CG with efficient data 
    augmentation algorithm.
%
%
%
In this paper, to further overcome the decoding 
    instability problem, {\bf we 
    propose a robust correspondence finding algorithm based on Markov random 
    field (MRF) optimization}.
In the method, the global correspondences for phase unwrapping is initially done by graph convolutional network (GCN), and then, MRF optimization is applied in the next step 
by voting scheme.
%
{\bf We also propose to refine the phase information using explicitly detected 
    laser curves}. Since inference of deep network may have some bias, it is 
    effectively refined by b-spline and Gaussian kernel based interpolation method. 
%
Experiments are conducted to show the effectiveness of the proposed method using 
real data, where captured images include strong noises and complex textures.

\section{Related works}
Active stereo 
is 3D shape measurement technique,
where a pattern is projected from a projector onto the object, and 
shapes are reconstructed by stereo algorithm~\cite{ActiveLight:Ikeuchi20:Springer}.
%
Among them, one-shot scan,
which uses a spatial-coded pattern to measure the shape by a single image, draw wide attention,
since it is suitable for measurement of moving objects~\cite{kawasaki2008dynamic, ulusoy2009one, Furukawa_2022_WACV, Kinect} or by moving cameras~\cite{Furukawa:embc20,Furukawa:embc2021}.
A key technique of one-shot scan is efficient correspondence finding between the projection pattern and the projected pattern on the object surface in the captured image~\cite{proesmans1997one,ulusoy2009one}.
Recently, various methods have been developed for detecting the correspondences in stereo pairs using deep learning~\cite{zagoruyko2015learning,vzbontar2016stereo, sagawa2011dense,furukawa2016shape}, however,
Achieving a dense measurement is also a challenging problem using a spatial-coded pattern~\cite{kawasaki2015active,sagawa2009dense}.
The proposed system projects a grid-like patterns and uses its graph structure to obtain correspondences using a graph convolutional network (GCN)~\cite{defferrard2016convolutional} and densify the correspondences by using phase information extracted by using U-Net.

\section{Overview and the proposed method} 
\begin{figure}[t]
\vspace{-0.5cm}
 \centering
 \includegraphics[width=0.8\linewidth, bb=0 0 755 442]{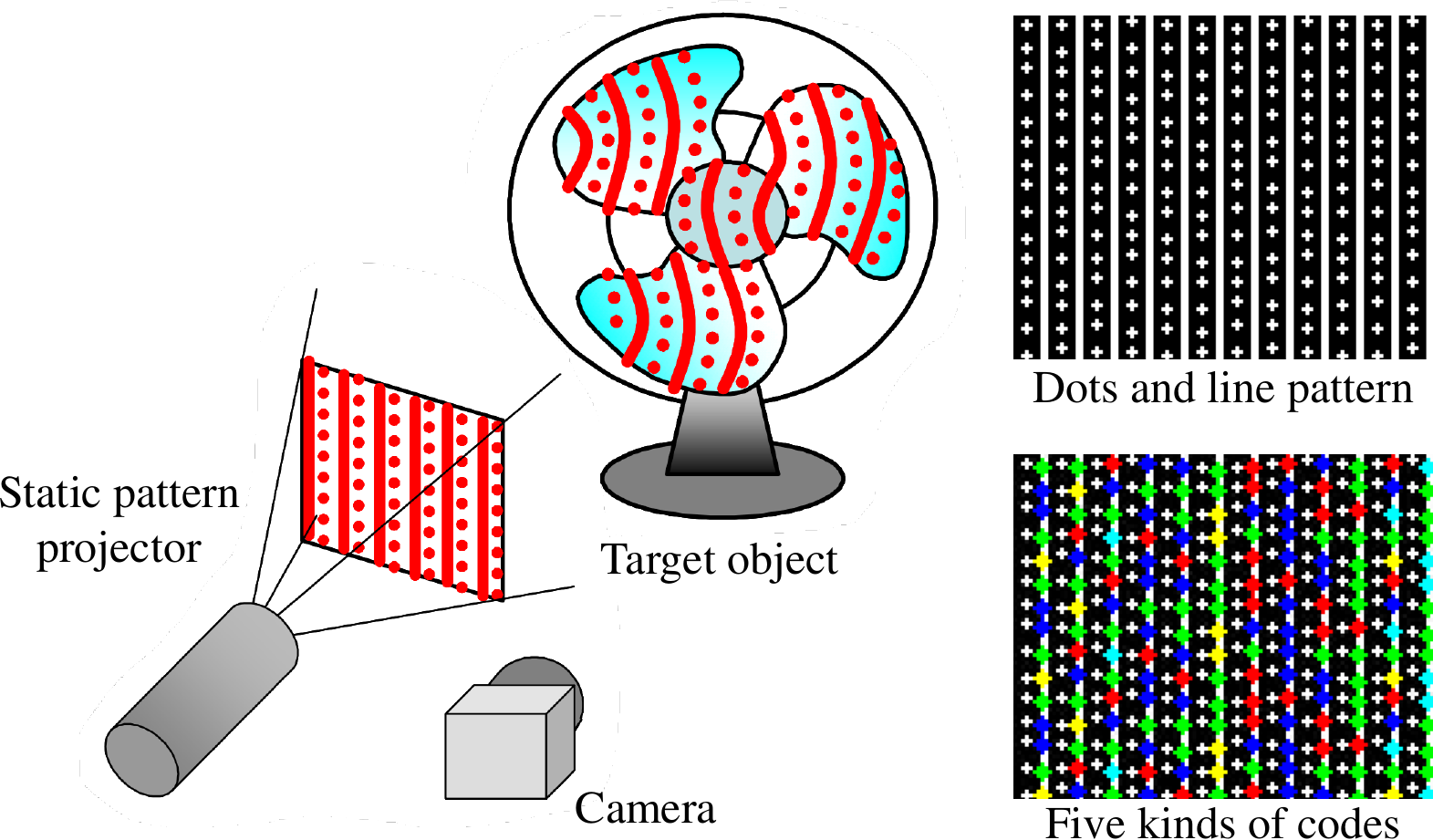}
\vspace{-0.1cm}
\caption{(Left) One-shot active stereo system, (Right top) our projection pattern and (Right bottom) embedded codes.~\cite{Furukawa_2022_WACV}
}
 \label{fig:pattern}
 \vspace{-9mm}
\end{figure}

For the shape measurement of the proposed method, we use common
one-shot 3D scanning setup as shown in \figref{fig:pattern}(left).
The system consists of a projector and a camera, and the projection pattern has a grid-like structure, whose nodes have five kinds of features, as shown in the bottom right of \figref{fig:pattern}.
%
As the correspondences are obtained from only single image, the feature points 
are sparse, which is a open problem for one-shot 3D scan. 
In the proposed method, 
we follow  dense shape measurement technique proposed by Furukawa~\cite{Furukawa_2022_WACV}.

\begin{figure*}[htb]
 \vspace{-3mm}
\begin{center}
\includegraphics[width=0.82\linewidth, bb=0 0 1128 351]{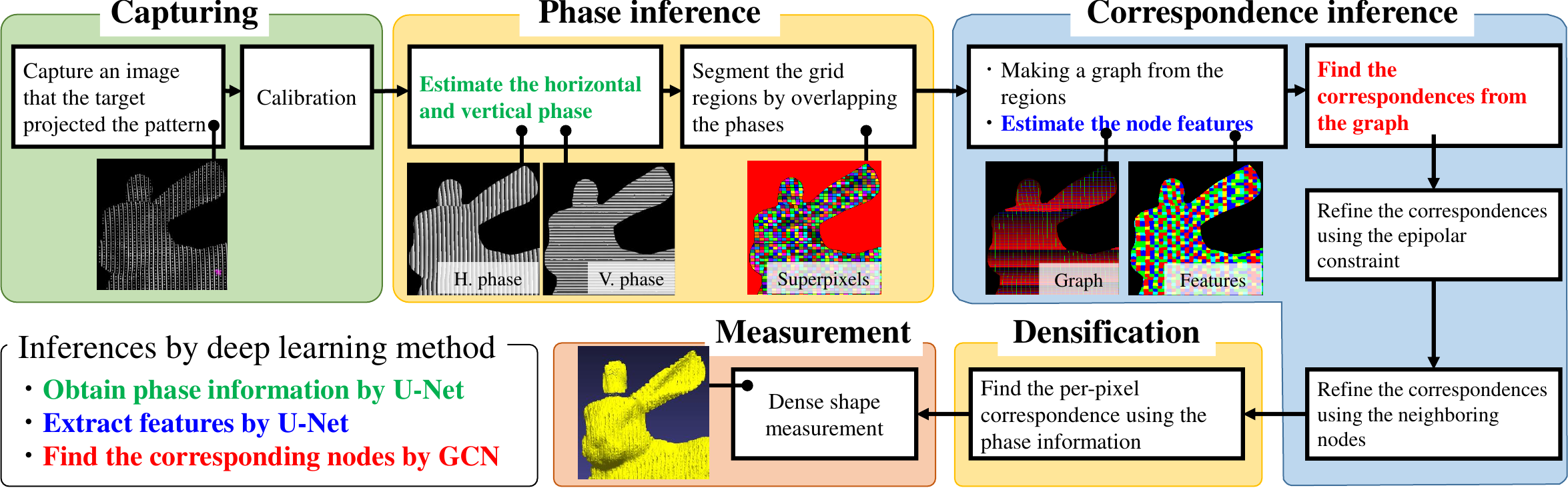}
\end{center}
\vspace{-7mm}
  \caption{Overview of the dense measurement}
\label{fig:overview}
\vspace{-3mm}
\end{figure*}

\figref{fig:overview} shows the overview of the system. 
The system achieves the dense reconstruction by the following two processes.
First, the sparsely distributing node correspondences between the projection pattern and the projected pattern in the input image are found by
%
graph convolutional network (GCN).
%
Then, dense correspondences are retrieved by interpolation 
algorithm using phase information between sparse nodes, where
the interpolation is achieved by 
U-Net, which extracts the phase information form the input image~\cite{Furukawa_2022_WACV}.
%
\ifx
In the method, since the system is based on a stereo system, epipolar 
constraints are further used to filtering nice correspondences.
However, due to the nature of the graph convolution, the corresponding nodes may be mistaken by nearby nodes.
Here, we refine the candidates of the corresponding node by using the information of the surrounding nodes.
Finally, 3) we obtain per-pixel corresponding using the phase information and corresponding nodes and reconstruct a dense 3D shape.
\fi


Since there are several problems with the system in the case that difficulty on 
preparing training data or input images include strong noises or the pattern is 
projected on a surface with textures, we propose fundamental improvements for the solution as explained in the following sections.

\subsection{Efficient pre-training by computer graphics}

The method measures arbitrarily shaped objects.
The proposed method uses computer graphics to generate the training data.
In order to deal with the arbitrary shapes, a pattern is projected onto a large number of object surface generated by computer graphics.
Furthermore, that allows us to flexibly use different patterns.
That is, the measurement can be applied without changing the algorithm when the projection pattern is changed, but only training data are changed instead.
%
In addition, we add noise to the training data, taking into consideration of the capturing condition, for example, the noise from sensors and speckle noise from laser projectors.
By adding these variation into the training data, robust estimation against noise can be achieved.

\subsection{Corresponding nodes refinement by solving MRF} 
One of the important factor for active stereo method is finding the correspondences between the projection and the projected patterns accurately.
The problem can be seen as an optimization problem that follows to Markov random field (MRF) 
as defined as follows.
\begin{eqnarray}
E(X) = \sum_{v\in V} g_{v}(X_{v}) + \sum_{(u, v)\in E} h_{uv}(X_{u}, X_{v}).
\end{eqnarray}
The first and second terms are the data and smoothness term, respectively.
We regard the data cost becomes smaller by the probability of the node estimation by GCN, while the smoothness cost becomes smaller when the connection on the pattern is correct.
This is because the GCN utilizes a multi-resolution analysis, the target node tend to be assigned to be positionally close node, resulting in a wrong correspondence.
Using the connections of the target node and the neighboring node, we accurately estimate the correspondence between the node in the projection pattern and projected pattern.

\begin{figure}[t]
 \vspace{-3mm}
 \centering
 \includegraphics[width=0.95\linewidth, bb=0 0 958 500]{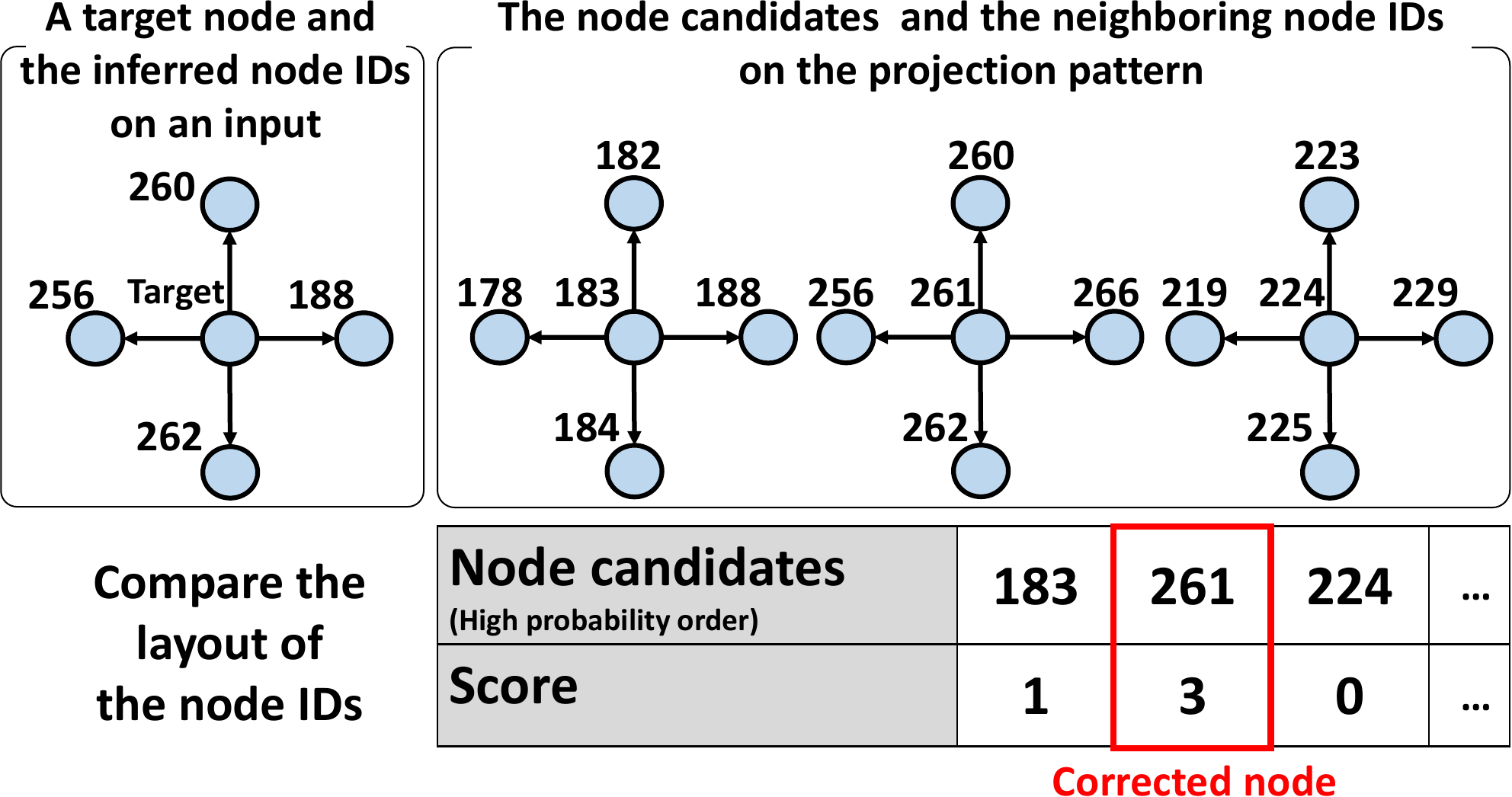}
 \vspace{-1mm}
 \caption{Node correction using the neighboring node IDs}
 \label{fig:search_node}
 \vspace{-5mm}
\end{figure}

To solve the MRF problem, several typical methods are known, such as belief propagation or graph cut, however it tends to be time-consuming~\cite{Veksler:cvpr2005,mrf2003}.
The computation time is one of the important factors for our one-shot active scanning system, therefore, we implemented a naive but computationally efficient method based on heuristics. 
In our method, we use a relationship between the target node and the inferred neighbors on the projection pattern.

Suppose the node inference is done correctly, the layout of the target node and the neighboring node in the input image should be 
the same as those on the projection pattern.
%
In \figref{fig:search_node}, circles represent nodes and the number represents the inferred node ID,
\ie, the detected node on the center, whose ID is 183, has the neighboring four nodes.
For this node, the corresponding candidate node IDs obtained by inference were 183, 261, 244, $\cdots$ in the order of high probability.
The scores in the table in \figref{fig:search_node} shows the number of the correct node IDs.
The score is 1 when the connection of the target node and the neighboring node on the input are the same on the projection pattern.
For example, the connection of the target node and right node (ID 188) is the same for node ID 183, while other connections are not, resulting in the score to be 1.
We compute the score for all the candidates (ID 183, 261, 244,$\cdots$) and
set the node ID having highest score, 
which is, in this case, 261.
Since there are errors in the inference process, we repeat the process until convergence to estimate the corresponding node IDs of all target nodes.

\subsection{Pixel-wise phase refinement by Gaussian kernel}
\label{sec:spline}


\begin{figure}[t]
 \vspace{-1mm}
\begin{center}
    \includegraphics[width=0.95\linewidth, bb=0 0 578 293]{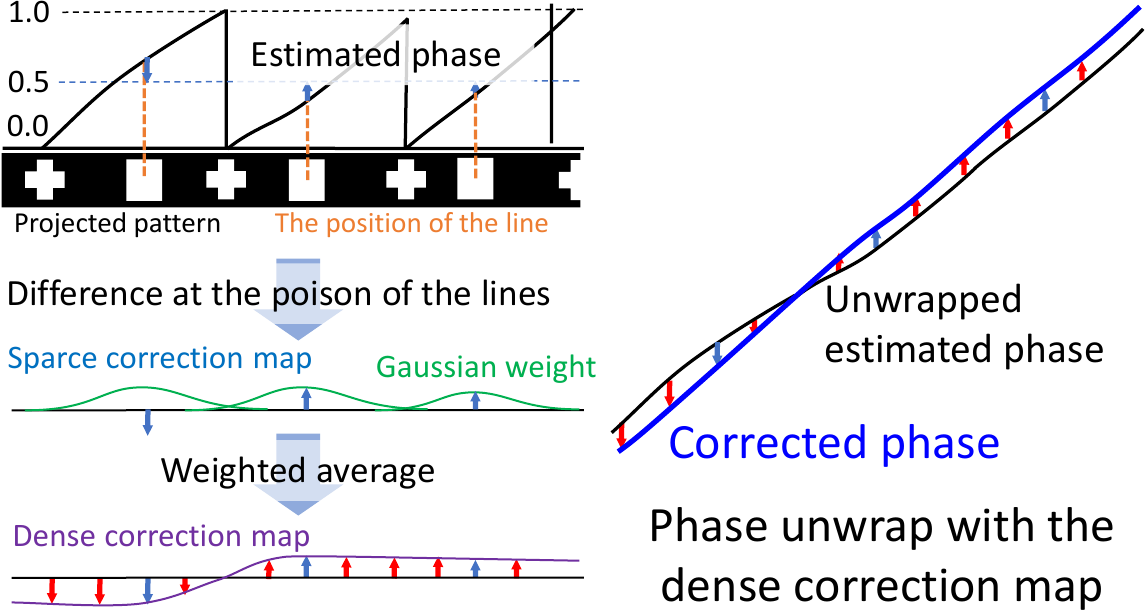}
\end{center}
 \vspace{-4mm}
  \caption{Phase correction using the position of the lines on projected pattern. Left) the process to obtain the dense correction map and Right) the corrected phase.}
\label{fig:phaseCorrection}
\vspace{-9mm}
\end{figure}

As the phase estimation by U-Net is spatially imitated by the sparse distribution of the dots on the projection pattern, 
it is hard to detect phases on a surface with high frequency.
Therefore, we propose a method that correct the phase estimation by lines in the pattern (\figref{fig:phaseCorrection}).
Ideal phases are a saw tooth and range from 0 to 1, starting from a node to next node.
The lines exist in the middle of the nodes and appear repeatedly on the pattern.
Therefore, ideally, the phase value on the line is 0.5.
We regard the difference of the phase value inferred by U-Net and 0.5 (ideal phase value on the line) as the correction value of the phase. 
We apply Gaussian filtering to the correction value based on our assumption that the distribution follows to Gaussian.
By using the dense correction map, we achieve a phase that close to the ideal phase.

\section{Evaluation by experiments}

\begin{figure}[t]
\vspace{-2mm}
 \centering
 \includegraphics[width=0.9\linewidth, bb=0 0 1527 1070]{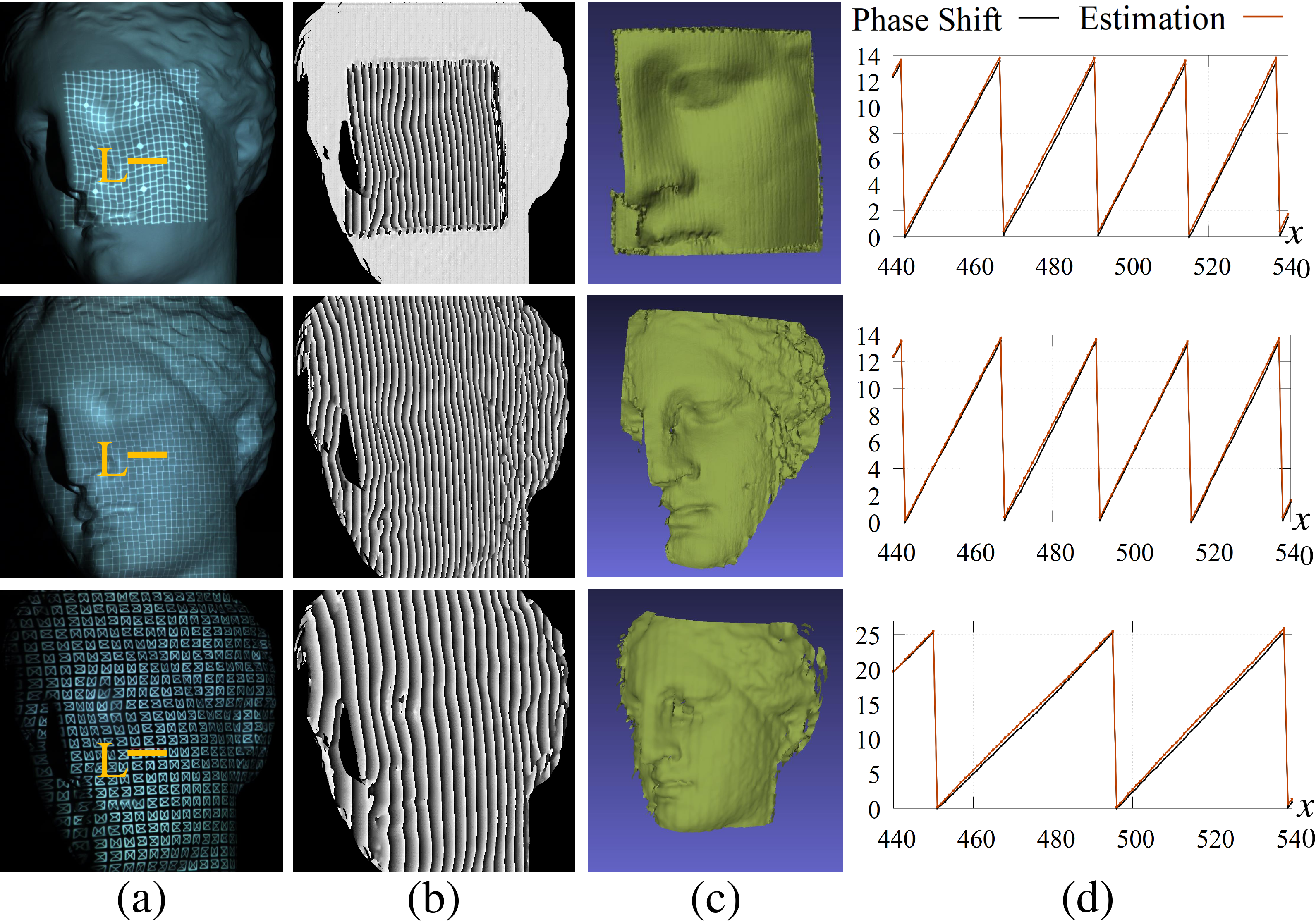}
 \vspace{-1mm}
\caption{Phase estimation by various grid-like patterns. (a) Input, (b) estimated phase values, (c) measurement results, and (d) The profile compared to the ground truth (phase shift method ~\cite{ikeuchiactive}) along with the lines L. The pattern used are from ~\cite{Furukawa:embc20} (the 1st row), an original pattern (the 2nd row), and ~\cite{jia2021one} (the 3rd row), respectively.  
}
 \label{fig:phaseDetection}
\vspace{-8mm}
\end{figure}

\subsection{The data augmentation by CG images}
By changing the settings of rendering parameters, computer graphics techniques are suitable for generating training data of several conditions.
We trained U-Net by applying the projection patterns from the methods~\cite{Furukawa:embc20, jia2021one} for the phase detection (\figref{fig:phaseDetection}).
\figref{fig:phaseDetection}(d) shows the comparison of the phase estimation with the ground truth obtained by~\cite{ikeuchiactive}. 
From the top to bottom row, the RMSEs are 0.3061, 0.2358, and 0.5164, respectively. 
The measurement results shows the phases are correctly estimated using the patterns. 
\begin{figure}[t]
 \centering
 \includegraphics[width=1.0\linewidth, bb=0 0 844 344]{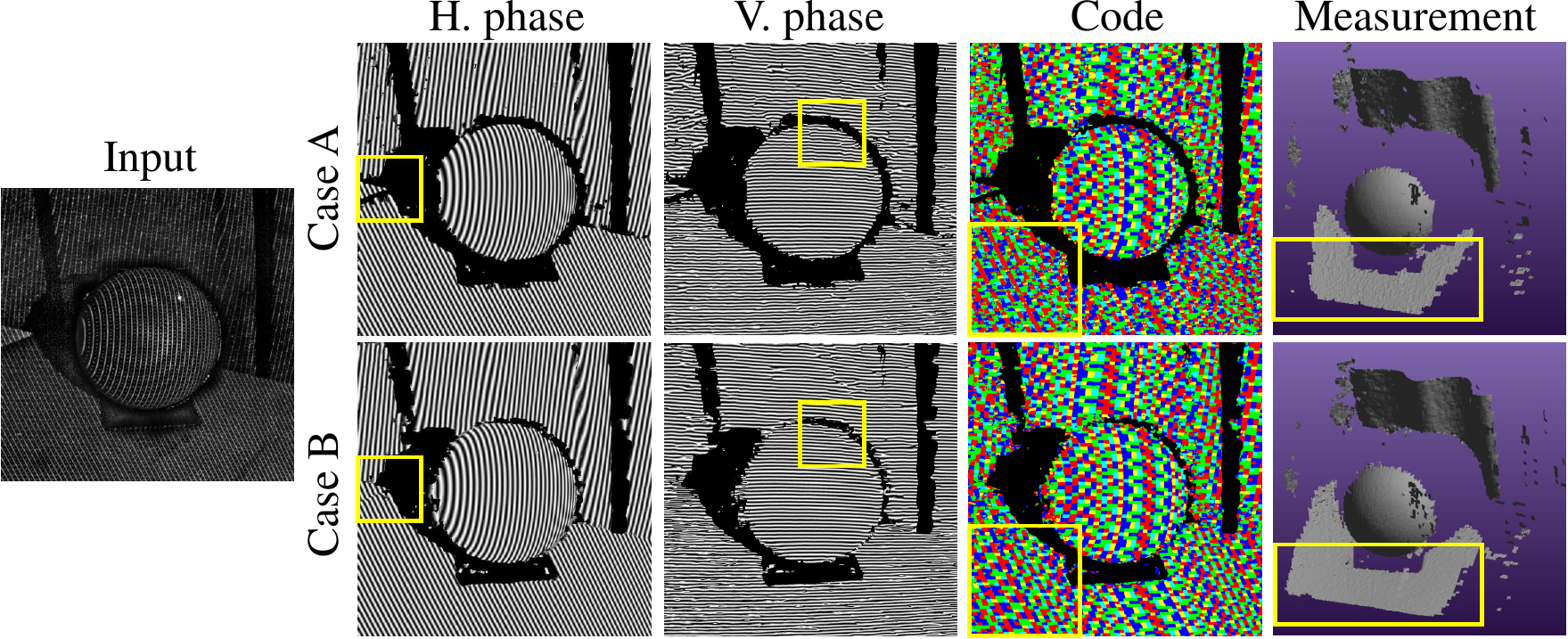}
 \vspace{-5mm}
\caption{Comparison of the performances by data augmentation.
From left to right, Input, the inferences horizontal phase, vertical phase, code, and the measurement result.
The top row shows the results using the training data with less noises (Case A), while the bottom with more noises (Case B).
}
 \label{fig:result_train_Unet}
\vspace{-3mm}
\end{figure}

Next, we compare the other estimation results using the training data with different noise parameters.
In Case A and B, we added noise that follows a Gaussian distribution.
Case B has more noise variations than Case A.
That is, the Gaussian distribution whose mean is set to 60 and standard deviation is 180.
We set the roll rotation of the camera to $0, \pm{2}, \pm{4}, \pm{6}$ and $\pm{8}^{\circ}$.
In our experience of trying various values, the best results were obtained when training at these values.
In addition, we modified the training data to have a variety of brightness.
\figref{fig:result_train_Unet} shows the visualization results of the horizontal, vertical phase estimation, code, and shape measurement result from the input. 
As it can be seen in the area surrounded by the yellow square, wider areas can be measured in Case B than Case A. 

\subsection{Evaluation of correspondence search based on MRF optimization}

\begin{figure}[tb]
 \vspace{-2mm}
 \centering
  \includegraphics[width=0.9\linewidth, bb=0 0 409.89 168.09]{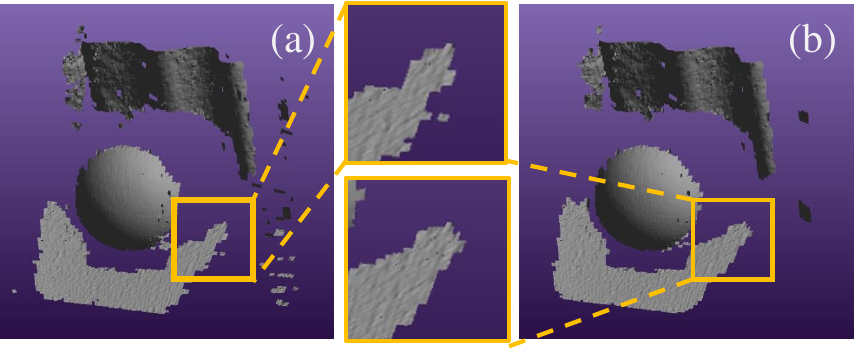}
 \vspace{-1mm}
\caption{
The measurement result (a) without the correspondence refinement, and (b) with the correspondence refinement (the proposed method).
}
 \label{fig:sarch}
  \vspace{-8mm}
\end{figure}

\figref{fig:sarch} shows the measurement results (a) before and (b) after the corresponding node correction.
From the figure, it is clearly shown that the proposed method efficiently reconstructs the part where the previous method cannot recover because of wrong correspondences.

\subsection{Evaluation of the phase refinement}
\begin{figure}[ht]
 \vspace{-5mm}
  \centering
  \begin{minipage}[b]{0.45\linewidth}
 \vspace{-15mm}
%
 \includegraphics[width=1.0\linewidth, bb=0 0 609 529]{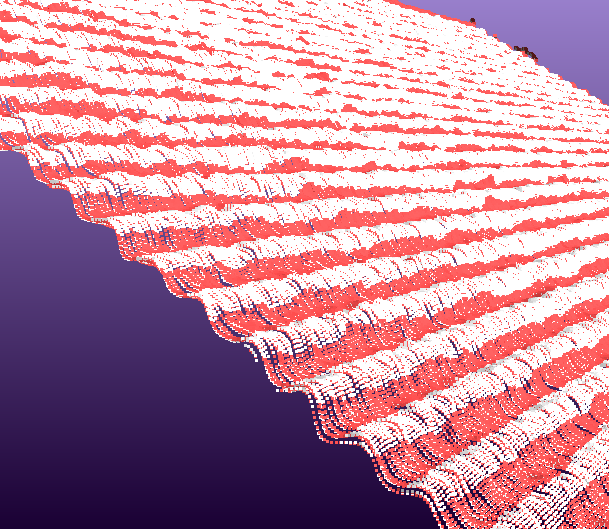}
  \end{minipage}
  \begin{minipage}[b]{0.45\linewidth}
 \includegraphics[width=1.0\linewidth, bb=0 0 609 526]{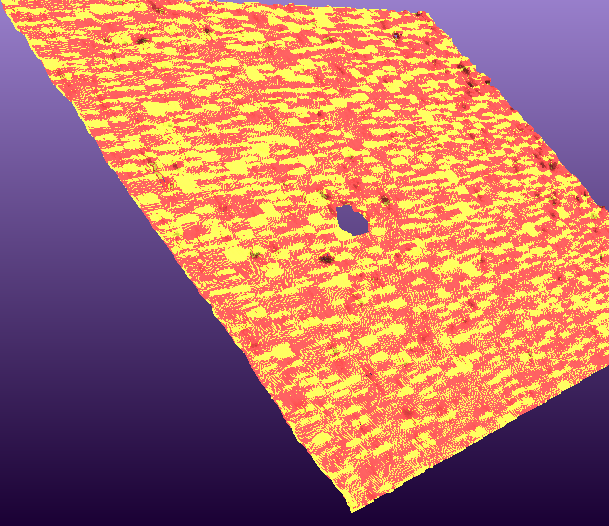}
 \end{minipage}
 \\(a) CG data \hspace{1cm} (b) real data
  \vspace{-0mm}
    \caption{Phase refinement results. (a) Red: before and white: after refinement. (b) Red: before and orange: after refinement. In (a), it is confirmed that white shape has recovered high frequency shapes.}
    \label{fig:phase_refinement}
\end{figure}

We applied our phase refinement method to two data set for evaluation.
The first one is CG generated high-frequency shape and the second is real data of planar board.
The results of RMSE is shown in \tabref{table:phaseCorrectionMeasurement}, where we can confirm that accuracy is improved by our method. Reconstructed shapes before and after refinement are shown in \figref{fig:phase_refinement}, where we can clearly confirm that high frequency shapes are correctly recovered by our method for CG data.

\begin{table}[ht]
  \begin{center}
  \begin{tabular}{lcr}
    \hline \hline
    RMSE (mm) & Before & After \\
    \hline 
    CG data    & 1.06 & 0.65\\
    real data  & 0.95 & 0.92 \\
    \hline
 \end{tabular}
  \vspace{+3mm}
  \caption{Pixel-wise phase refinement by Gaussian kernel.}
 \label{table:phaseCorrectionMeasurement}
 \vspace{-12mm}
 \end{center}
\end{table}

\subsection{Shape measurement results}
\begin{figure}[t]
 \vspace{-4mm}
\begin{center}
 \includegraphics[width=0.9\linewidth, bb=0 0 1335 922]{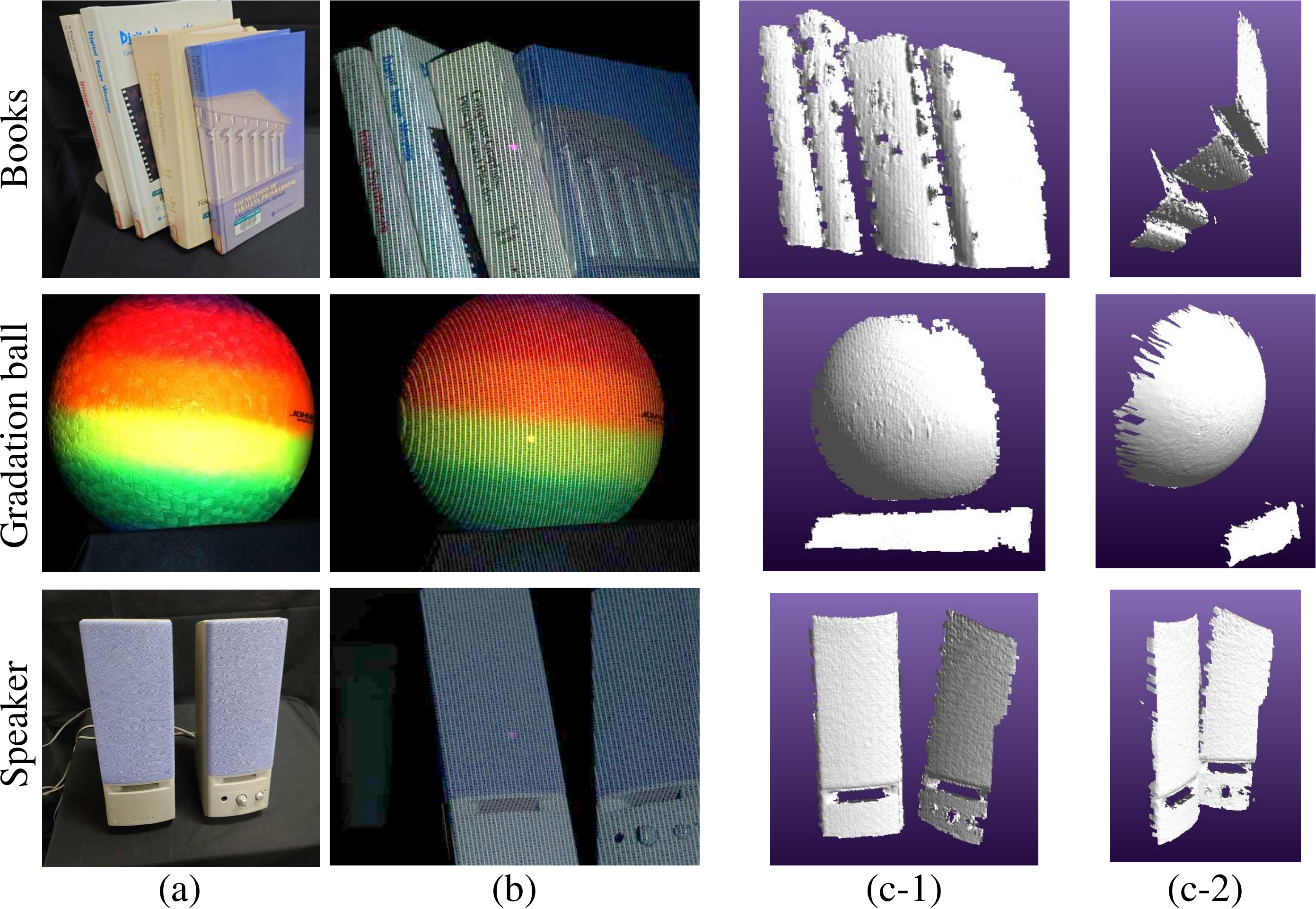}
 \vspace{-1mm}
    \caption{
    The dense shape measurement result of the proposed method.
    We applied our method to the objects whose surface have textures.
    (a) the appearance of the target objects, (b) the captured images, and (c) the measurement results seen from different views.
    }
\label{fig:3dreconst_Dotline}
\end{center}
\vspace{-8mm}
\end{figure}

Finally, we show the results of dense measurement using the projection pattern shown in \figref{fig:3dreconst_Dotline}. 
Our method achieved the robust measurement to the objects that have large occlusions and spatially-varying textures.
As for the calculation time of the reconstruction, it is taking 20 to 30 seconds per image.

\section{Conclusions}
In this paper, we propose a method to improve the accuracy of shape measurement 
of one-shot 3D scan.
In order to improve the accuracy 
we first augmented the training data by computer graphics by considering the situation of image captures and realistic noises.
The inference of correspondence nodes between patterns by the GCN is prone to wrong with neighboring nodes due to its multi-resolution nature.
To solve this problem, we proposed a method to refine the inference by using the neighboring node information
between the projection pattern and projected pattern based on MRF optimization.
Finally, the proposed phase correction method make it possible to measure the details of target objects.
In the future, real-time system with high fps using deep neural network is planned.

\vspace{-0.2cm}
\section*{Acknowledgment}
\vspace{-0.3cm}
This work was supported by JSPS/KAKENHI JP20H00611, JP21H01457 and JP23H03439 in Japan.
\vspace{-0.2cm}



{\small
\bibliographystyle{IEEEbib}
\bibliography{shortSTRING,JabRef-cvglab,JabRef-cvglab2,cvglab, pattern}
}

\end{document}